\documentclass[final,3p,twocolumn,times]{elsarticle}

\usepackage{amsthm,amssymb,mathrsfs,mathtools,amsmath,algorithm}
\usepackage{algpseudocode} 
\usepackage[hidelinks]{hyperref}
\usepackage[dvipsnames]{xcolor}
\usepackage{acronym}
\usepackage{picins}

\renewcommand\vec[1]{\overrightarrow{#1}}
\newcommand\cev[1]{\overleftarrow{#1}}

\newcommand{\pd}[2]{\frac{\partial #1}{\partial #2}}
\newcommand{\eq}[1]{(\ref{#1})}
\newcommand{\fig}[1]{Fig.~\ref{#1}}
\newcommand{\sct}[1]{Section~\ref{#1}}

\newcommand{\mx}{\bf{X}}

\newcommand{\mw}{\bf{W}}
\newcommand{\mz}{\bf{Z}}
\newcommand{\mt}{\bf{T}}

\newcommand{\vx}{\bf{x}}

\newcommand{\vz}{\bf{z}}
\newcommand{\vt}{\bf{t}}
\newcommand{\ve}{\bf{e}}
\newcommand{\vvt}{\tilde{\bf{t}}} 
\newcommand{\vve}{\tilde{\bf{e}}} 
\newcommand{\vme}{\tilde{\bf{E}}} 

\newcommand{\DX}{\bf{\cev{\Delta} X}}

\newcommand{\DW}{\bf{\cev{\Delta} W}}
\newcommand{\DZ}{\bf{\cev{\Delta} Z}}

\newcommand{\Dx}{\bf{\cev{\Delta} x}}

\newcommand{\Dz}{\bf{\cev{\Delta} z}}

\newcommand{\vDz}{\bf{\cev{\Delta} \tilde{z}}}

\newcommand{\dx}{{\cev{\Delta} x}}

\newcommand{\dw}{{\cev{\Delta} w}}
\newcommand{\dz}{{\cev{\Delta} z}}

\newcommand{\fDW}{\bf{\vec{\Delta} W}}

\newcommand{\fDE}{\bf{\vec{\Delta} E}}

\newcommand{\fDz}{\bf{\vec{\Delta} z}}
\newcommand{\fDe}{\bf{\vec{\Delta} e}}

\newcommand{\fvDZ}{\bf{\vec{\Delta} \tilde{Z}}}
\newcommand{\fvDE}{\bf{\vec{\Delta} \tilde{E}}}

\newcommand{\fdw}{{\vec{\Delta} w}}

\newcommand{\loss}{\mathcal{L}}

\newtheorem{prop}{Proposition}

\acrodef{apa}[APA]{affine projection algorithm}
\acrodef{sgd}[SGD]{stochastic gradient descent}
\acrodef{bp}[BP]{backpropagation}
\acrodef{dnn}[DNN]{deep neural network}
\acrodef{ics}[ICS]{internal covariate shift}
\acrodef{selu}[SELU]{scaled exponential linear unit}
\acrodef{relu}[ReLU]{rectified linear unit}
\acrodef{svd}[SVD]{singular value decomposition}
\acrodef{mse}[MSE]{mean squared error}
\acrodef{mlp}[MLP]{multi-layer perceptron}
\acrodef{rnn}[RNN]{recurrent neural networks}

\makeatletter
\def\ps@pprintTitle{
  \let\@oddhead\@empty
  \let\@evenhead\@empty
  \def\@oddfoot{}
  \let\@evenfoot\@oddfoot}
\makeatother

\begin{document}

\begin{frontmatter}

\author[1]{Naeem Paeedeh}
\ead{naeim.p@aut.ac.com}

\author[2]{Kamaledin Ghiasi-Shirazi\corref{cor1}}
\cortext[cor1]{Corresponding author.}
\ead{k.ghiasi@um.ac.ir}

\address[1]{Department of Mathematics and Computer Science, 
Amirkabir University of Technology, Tehran, Iran.}

\address[2]{Department of Computer Engineering, Ferdowsi
University of Mashhad (FUM), Office No.: BC-123, Azadi Sq.,
Mashhad, Khorasan Razavi, Iran.}

\title{Improving the Backpropagation Algorithm with Consequentialism Weight Updates over Mini-Batches}

\begin{abstract}
Many attempts took place to improve the adaptive filters that can also be useful to improve \ac{bp}. Normalized least mean squares (NLMS) is one of the most successful algorithms derived from Least mean squares (LMS). However, its extension to multi-layer neural networks has not happened before. Here, we first show that it is possible to consider a multi-layer neural network as a stack of adaptive filters. Additionally, we introduce more comprehensible interpretations of NLMS than the complicated geometric interpretation in \ac{apa} for a single fully-connected (FC) layer that can easily be generalized to, for instance, convolutional neural networks and also works better with mini-batch training. With this new viewpoint, we introduce a better algorithm by predicting then emending the adverse consequences of the actions that take place in \ac{bp} even before they happen. Finally, the proposed method is compatible with \ac{sgd} and applicable to momentum-based derivatives such as RMSProp, Adam, and NAG. Our experiments show the usefulness of our algorithm in the training of deep neural networks.
\end{abstract}

\end{frontmatter}
\section{Introduction \label{sec:introduction}}
\ac{bp} \cite{rumelhart_learning_1986, bishop2006pattern} is a prevalent method of training neural networks \cite{LeCun2015DeepL}. The training involves many matrix operations. We know that regardless of GPU parallelization of matrix operations, forward and backward calculations are sequential. For example, in \cite{Jaderberg2017DecoupledNI}, they tried to decouple layers and use \ac{rnn}s to predict the gradients and update the weights asynchronously instead of waiting to get the actual gradients in the computation graph. Consequently, it is desirable to train deep neural networks in fewer iterations that means higher convergence rates, and it is better to use the wasted resources to do some computations asynchronously.\par
Another significant subject is the accurately chosen probability distribution of each part of the neural networks at the beginning of the training. Some issues may arise if something violates this assumption during the training. The training performance may suffer if some changes happen during the training, for example, when someone wants to modify the distribution of inputs or weights, injecting noise, or applying a new regularization technique that may invalidate the mentioned assumption. Therefore, a robust method of training can help the researchers to focus on new techniques instead of spending time to troubleshoot a network or tuning the hyper-parameters.\par
One way to achieve higher convergence rates is to use second-order methods instead of commonly used first-order derivatives in \ac{bp}. Newton and Quasi-Newton methods such as conjugate gradient \cite[pp. 101-102]{nocedal_numerical_2006} and L-BFGS \cite[pp. 177-180]{nocedal_numerical_2006} are examples of second-order methods or approximations of them to train in fewer iterations at the cost of much more extra calculations per iteration.\par
There are some other computationally cheaper methods to make \ac{bp} faster and more robust, such as introducing momentum terms \cite{rumelhart_learning_1986} or batch-normalization (BN) layer \cite{Ioffe2015BatchNA}. We want to find out why these approaches were successful and how to cultivate them. Before that, we should mention some aspects of \ac{sgd}.\par
The learning curves of the \ac{sgd} algorithm usually contain many fluctuations. This phenomenon may happen due to optimization over a small subset of input vectors or mini-batches at each iteration, which gives us a rough estimation \cite{LeCun2015DeepL}. Besides, plain \ac{sgd} shows weakness in some parts of the loss surface that bends sharply in one dimension than the other \cite[p. 296]{Goodfellow-et-al-2016}\cite{ruder2016overview}. Therefore, we should choose the lower learning rates that do not cause overshooting. We explain it further in \sct{experiments}.\par
In the following, we explain how momentum and BN methods make it possible to choose higher learning rates. Next, we explain how NLMS is related to this subject.\par
The momentum method \cite{rumelhart_learning_1986} and its extensions such as Adam, NAG \cite{ruder2016overview}, Adagrad \cite{Duchi2011AdaptiveSM}, and Adadelta \cite{Zeiler2012ADADELTAAA} help to stabilize and speed-up the training by bringing some eigen components of the system closer to critical damping \cite{Qian1999OnTM}. These methods reduce the oscillations of parameters that change a lot around their optimum values, for example, by guiding them to move straight towards their optimum values. Besides, the momentum helps speeding up the convergence of the other parameters that slowly move towards their optimum values. Therefore, we can see that the momentum method makes the behavior of the parameters more predictable than before. Less chaotic behavior of previous layers means higher learning rates can be applied, which means faster training.\par
According to \cite{Ioffe2015BatchNA}, BN helps to train neural networks without the concern of changes that may happen to the probability distribution of the inputs of each layer that they assume it as a source of unpredictability during the training, which slows down the training by forcing one to pick lower learning rates and to initialize the parameters carefully. It normalizes individual features of the outputs by using the first two moments by the statistics of each mini-batch. Besides, it introduces two new learnable variables to restore the representation power of the network by finding the preferred scale and shift over the training. These initiatives enabled them to use higher learning rates to achieve state-of-the-art accuracy with 14 times fewer training steps.\par
The success of BN has at least two main reasons: Firstly, it makes the outputs of each layer, hence the behavior of the inputs of the next layers, more predictable. The possibility of using higher learning rates for the modified architecture is an effect of the mentioned predictability, not a cause. The next one that gets less attention is that BN adds more learnable parameters that also increase the power of a network. In \cite{frankle2020training}, they confirm that just BN parameters are enough to train a neural network while freezing the other parameters.\par
LMS algorithm is the common origin of neural networks and adaptive filters \cite{widrow_30_1990, widrow_adaptive_1960}. While research on neural networks has been targeting multi-layer architectures, research in the field of adaptive filters has been focused on single-layer architectures mainly for signal processing. In neural networks, \ac{bp} over mini-batches is the core method for training multi-layer networks, and any improvement to this algorithm is of great practical importance. In this paper, we show that training individual layers with LMS is equivalent to training with \ac{bp}. Then, we try to transfer the improvements to the LMS algorithm in the field of adaptive filters to the training of \ac{dnn}s.\par
LMS algorithm trains adaptive filters in an online fashion by reducing the errors through an iterative process. An adaptive linear element (Adaline) network \cite{widrow_30_1990,widrow_adaptive_1960} is a single-layer neural network with a linear activation function. Subsequent investigations revealed that it is hard, if not impossible, to pick a learning rate to make the convergence faster while maintaining stability at the same time that is a result of its sensitiveness to the norm of the input vector \cite{widrow_30_1990}. Besides, the filter learns the stronger inputs faster, while the weaker inputs need more steps to train. Therefore, we expect a kind of bias towards stronger inputs. Also, it may easily be confused when suddenly encounters an outlier by jumping over other parts of the loss surface far from the current point, hence forgetting the earlier learned parameters.\par
NLMS \cite[pp. 333-337]{haykin2014adaptive} solves the mentioned problems. It is a well-known method in regression tasks, and it is also known as $\alpha$-LMS \cite{widrow_30_1990, hassoun1995fundamentals} in classification variants. Although it is accepted as a normalization method as the name suggests, we show it does not normalize the input by adjusting the learning rate, but rather it divides its unit vector by its norm that means it divides the input vector by the square of its norm, not the norm.\par
Two questions need to be answered to extend the success of NLMS to \ac{dnn}s. First, how someone can generalize the underlying idea to a network with multiple layers? The second one arose when we extended it to a \acf{mlp} and observed instabilities during the mini-batch training. The second question is, what was the cause of instability? To address the first issue, we show that each layer follows some virtual targets. We prove that optimization with \ac{bp} is equivalent to applying the LMS algorithm on independent layers. We solve the second problem by considering how each sample of the input deflects another input sample from pursuing its virtual target. This phenomenon makes the outputs of each layer unpredictable, which makes the inputs of the next layer unstable.\par
It is important to pinpoint that in practice, some groups of neurons, not just a single neuron, determine the outcome. For instance, networks that do not rely on individual neurons exhibit better generalization \cite{morcos2018importance}. In fact, instead of just considering a group of scalar inputs as vectors in on-line training, one should consider those vectors as each part of a matrix of a mini-batch \cite{LeCun2015DeepL}.\par
Ozeki and Umeda improved NLMS by considering a geometric interpretation of a batch of last seen inputs \cite{ozeki1984adaptive} \cite[pp. 345-350]{haykin2014adaptive}. This viewpoint has some limitations. For instance, it is hard to generalize. Since our proposed method does not suffer from such complications, we easily extend it to the convolution operator.\par
Finally, Atiya and Parlos put constraints on gradients w.r.t. the outputs to train the recurrent neural networks \cite{atiya_new_2000}. In their method, they considered the states as the control variables. The weight modifications were elicited from the changes in the states. As we show in \sct{sec:past works}, NLMS can also be explained by the gradients w.r.t. the outputs of the neurons instead of the weights.\par
The paper proceeds as follows: First of all, we introduce notations in \sct{sec:notations}. Second, some past works are analyzed in \sct{sec:past works} as the basis for the next section. After that, we propose our method in \sct{proposed method}. Next, we investigate the effectiveness of the algorithm with some experiments in \sct{experiments}. At last, the paper concludes in \sct{sec:conclusion}.
\section{Notations \label{sec:notations}}
In this section, we provide a summary of the notations that we use in this paper. Bold letters indicate vectors, and bold capital letters represent matrices. Lower case letters denote the matrix and vector elements. For instance, $m_{ij}$ refers to the $i$th row and $j$th column of the matrix $\bf{M}$.\par
To explain other notations, consider a feed-forward network with $L$ layers (or $L-1$ hidden layers) that we want to train with a batch size of $N$. $D^{(\ell)}$ denotes the dimension of the output for layer $\ell \in \{0, 1, \dots, L\}$. As two special cases, $D^{(0)}$ is the dimension of the inputs of the network and $D^{(L)}$ is the number of classes in a typical classification task.\par
\begin{figure}[!t]
  \centering \includegraphics[width=3in]{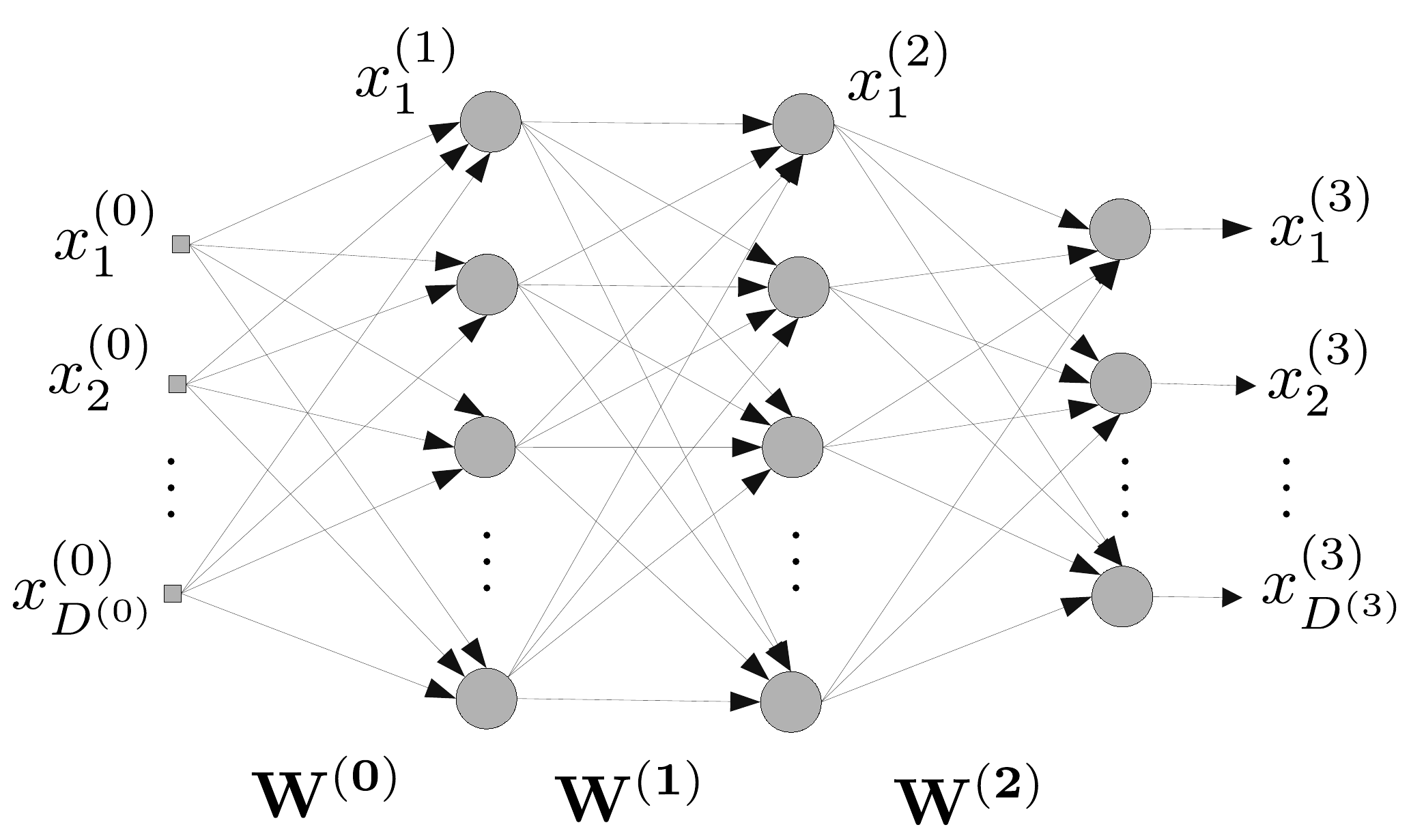}
  \caption{An \ac{mlp} network with 3 layers (2 hidden layers).}
  \label{fig:mlp architecture}
\end{figure}

For layer $\ell \in \{0, 1, \dots, L\}$, we refer to the input matrix by $\mx^{(\ell)}$, which is $D^{(\ell)} \times N$, and for layer $\ell \in \{0, 1, \dots, L-1\}$, we refer to the weight matrix by $\mw^{(\ell)}$, which is $D^{(\ell+1)} \times D^{(\ell)}$. These notations are applied in a 3-layer \ac{mlp} network in \fig{fig:mlp architecture} as an example.\par
In the forward pass, the outputs of the layer $\ell$ are calculated by
\begin{equation}
  {\mz}^{(\ell)} = {\mw}^{(\ell)} \mx^{(\ell)}. \label{forward formula with step}
\end{equation}
These outputs then pass through a non-linearity or activation function $\phi$ to create the inputs of the next layer as
\begin{equation}
  \mx^{(\ell+1)}=\phi(\mz^{(\ell)}).
\end{equation}
In the final step, the loss function $\loss$ finds the discrepancy between the outputs of the network ($\mx^{(L)}$) and desired targets, which are referred to by the matrix of $\bf{T}$. As a special case, $\loss^{\text{MSE}}$ refers to the \ac{mse} loss function. Henceforth, we omit the layer number when we discuss a known specific layer to keep formulas simple.\par
Finally, $\cev{\Delta}$ denotes the gradients in the backward pass and $\vec{\Delta}$ denotes our expected changes in the forward pass of the current mini-batch. For example, $\DW$ is the matrix of gradients of the loss w.r.t. the weights, and $\fDW$ is the matrix of actual changes to make to the weights by $\mu$ as the learning rate coefficient. We also use $:=$ for definitions or assignments.
\section{Background \label{sec:past works}}
In this section, we briefly review BP, LMS, and NLMS algorithms. This background knowledge is needed to introduce our consequentialism idea that we want to extend to mini-batch training of the BP.\par
\subsection{Backpropagation \label{BP}}
\ac{bp} assigns the needed correction to each weight in a neural network by following the long path to it by the chain rule \cite{haykin2009neural, schmidhuber2015deep}. Besides, it utilizes dynamic programming to store the results of the intermediate calculations to prevent redundant recalculations.\par
\ac{bp} can be explained by mathematical induction \cite{bishop2006pattern, haykin2009neural}. Algorithm \ref{alg:forward} shows the forward phase for a feed-forward network in the mini-batch mode. Here, we see this weight update rule is calculable for all layers. In the next section, we explain why and how one can deal with each layer as an adaptive filter.
\begin{algorithm}[H]
  \caption{Forward phase for a feed-forward network}
  \label{alg:forward}
  \begin{algorithmic}[1]
    \Require input $\mx^{(0)}$, an array of weight matrices $\mw$,
    target matrix $\mt$, number of layers $L$, and activation
    function $\phi$ \For{$\ell = 0, 1, \dots, L-1$} \State
    $\mz^{(\ell)} \gets \mw^{(\ell)} \mx^{(\ell)}$ \State
    $\mx^{(\ell+1)} \gets \phi(\mz^{(\ell)})$ \EndFor \State Compute
    Loss $= \mathcal{L}(\mx^{(L)}, \mt)$ \State \Return
    $\mx,\mz,\text{Loss}$
  \end{algorithmic}
\end{algorithm}
\begin{prop} \label{thm:Backpropagation}
    The weight update rule of
      \begin{equation}
        \fDW^{(\ell)} := -\mu \DZ^{(\ell)} {\mx^{(\ell)}}^T, \quad 0 \le \ell \le L-1
      \end{equation}
  is correct and calculable for all layers.
\end{prop}
\proof
First, we show that one can calculate $\DX^{(L)}$ and $\fDW^{(L-1)}$ for the last hidden layer. Then, we prove that by having $\DX^{(\ell)}$ one can calculate $\fDW^{(\ell-1)}$ and $\DX^{(\ell-1)}$ for the previous layer.\par
To begin, we find the derivatives of the loss w.r.t. the outputs of the network as follow:

\begin{equation}
  \dx_i^{(L)} = \pd{\mathcal{L}(\vx^{(L)}, \vt)}{x_i^{(L)}},
\end{equation}
where $i \in \{1, 2, \dots, D^{(L)} \}$, \(\dx_i^{(L)}\) is the gradient of the loss w.r.t. the \(x_i^{(L)}\). It can be shown in the compact form of
\begin{equation}
  \Dx^{(L)} = \pd{\mathcal{L}(\vx^{(L)}, \vt)}{\vx^{(L)}}.
\end{equation}
By applying the chain rule again to the activation function of the last layer, we have:
\begin{align}
  \dz_i^{(L-1)} &= \pd{\mathcal{L}(\vx^{(L)}, \vt)}{z_{i}^{(L)}}
                  =\pd{\mathcal{L}(\vx^{(L)}, \vt)}{x_{i}^{(L)}}
                  \pd{x_{i}^{(L)}}{z_{i}^{(L-1)}}\\
                &= \dx_i^{(L)} \phi^\prime (z_i^{(L-1)})
\end{align}
or
\begin{align}
  \Dz^{(L-1)} &= \Dx^{(L)} \circ \phi^\prime (\vz^{(L-1)}),
\end{align}
where $\circ$ denotes the element-wise product (Hadamard product). Next, the weight parameters should be changed in the opposite direction of the gradients. Consequently, for $i \in \{1, 2, \dots, D^{(L)}\}$, $j \in \{1, 2, \dots, D^{(L-1)}\}$ we have:
\begin{align}
  \fdw_{ij}^{(L-1)} &:= -\mu \dw_{ij}^{(L-1)}
                      = -\mu \pd{\mathcal{L}(\vx^{(L)}, \vt)}{w_{ij}^{(L-1)}}\\
                    &=  -\mu \pd{\mathcal{L}(\vx^{(L)}, \vt)}{x_{i}^{(L)}} \pd{x_{i}^{(L)}}{z_{i}^{(L-1)}}
                      \pd{z_{i}^{(L-1)}}{w_{ij}^{(L-1)}} \\
                    &= -\mu \dx_i^{(L)} \phi^\prime (z_i^{(L-1)})
                      x_{j}^{(L-1)},
\end{align}
which can be shown in the compact form of
\begin{align}
  \fDW^{(L-1)} &:= -\mu \DW^{(L-1)} = -\mu \pd{\mathcal{L}(\vx^{(L)}, \vt)}{\mw^{(L-1)}}\\
               &= -\mu \Dx^{(L)} \circ \phi^\prime (\vz^{(L-1)}) {\vx^{(L-1)}}^T\\
               &= -\mu \Dz^{(L-1)} {\vx^{(L-1)}}^T. \label{eq:weight corrections in BP for the last layer}
\end{align}
For a mini-batch, the final weight changes are being calculated by summation of this individual weight corrections \cite{haykin2009neural} of \eq{eq:weight corrections in BP for the last layer} as
\begin{align}
  \fDW^{(L-1)} &:= \sum_{n=1}^N -\mu \Dz_{(n)}^{(L-1)}
                 {\vx_{(n)}^{(L-1)}}^T
                = -\mu \DZ^{(L-1)}
                 {\mx^{(L-1)}}^T,
\end{align}
where \(N\) is the batch-size, \(\Dz_{(n)}\) is the \(n\)th sample in the matrix of the gradients of the outputs, and \(\vx_{(n)}\) is the \(n\)th sample in the input matrix. Here, the base case is proved.\par
Now, we want to prove that by having $\DX^{(\ell)}$, one can calculate $\fDW^{(\ell-1)}$ and $\DX^{(\ell-1)}$. According to the chain rule, for $i \in \{1, 2, \dots, D^{(\ell)}\}$ and $j \in \{1, 2, \dots, D^{(\ell-1)}\}$ we have:

\begin{align}
  \fdw_{ij}^{(\ell-1)} &:= -\mu \dw_{ij}^{(\ell-1)}
                      = -\mu \pd{\mathcal{L}(\vx^{(L)}, \vt)}{w_{ij}^{(\ell-1)}}\\
                    &=  -\mu \pd{\mathcal{L}(\vx^{(L)}, \vt)}{x_{i}^{(\ell)}} \pd{x_{i}^{(\ell)}}{z_{i}^{(\ell-1)}}
                      \pd{z_{i}^{(\ell-1)}}{w_{ij}^{(\ell-1)}} \\
                    &= -\mu \dx_i^{(\ell)} \phi^\prime (z_i^{(\ell-1)})
                      x_{j}^{(\ell-1)} \Rightarrow\\
       \fDW^{(\ell-1)} &:= -\mu \DW^{(\ell-1)} = -\mu \pd{\mathcal{L}(\vx^{(L)}, \vt)}{\mw^{(\ell-1)}}\\
                    &= -\mu (\Dx^{(\ell)} \circ \phi(\vz^{(\ell-1)})) {\vx^{(\ell-1)}}^T\\
                    &= -\mu \Dz^{(\ell-1)} {\vx^{(\ell-1)}}^T. \label{eq:weight corrections in BP for arbitrary layer}
\end{align}
Again for a mini-batch, we sum up the individual weight corrections as
\begin{align}
  \fDW^{(\ell-1)} &:= \sum_{n=1}^N -\mu (\Dx^{(\ell)} \circ \phi(\vz^{(\ell-1)})) {\vx_{(n)}^{(\ell-1)}}^T\\
                  &= -\mu (\DX^{(\ell)} \circ \phi(\mz^{(\ell-1)}))
                 {\mx^{(\ell-1)}}^T\\
                  &= -\mu \DZ^{(\ell-1)}
                  {\mx^{(\ell-1)}}^T.
\end{align}
For the gradients of the inputs of the previous layer we have:
\begin{align}
  \dx_{j}^{(\ell-1)} &= \sum_{i=1}^{D^{(L)}} \pd{\mathcal{L}(\vx^{(L)}, \vt)}{x_{i}^{(\ell)}} \pd{x_{i}^{(\ell)}}{z_{i}^{(\ell-1)}} \pd{z_{i}^{(\ell-1)}}{x_{j}^{(\ell-1)}} \Rightarrow\\
                     &= \sum_{i=1}^{D^{(\ell)}} \dx_i^{(\ell)} \phi^\prime (z_i^{(\ell-1)})
                    w_{ij}^{(L-1)} \Rightarrow\\
  \Dx^{(\ell-1)} &= {\mw^{(\ell-1)}}^T (\Dx^{(\ell)} \circ \phi(\vz^{(\ell-1)})) =  {\mw^{(\ell-1)}}^T \Dz^{(\ell-1)}.
\end{align}
These are the gradients w.r.t. the elements of $\Dx$ that is needed for the bottom layer. Here, the induction step is also proved.
\endproof
To conclude, the backward phase can be simplified to
\begin{equation}
  \label{eq:recursive backward}
  \begin{cases}
    \DX^{(\ell)} := \pd{\mathcal{L}(\mx^{(\ell)}, \bf{T})}{\mx^{(\ell)}}, & \ell=L\\
    \DZ^{(\ell)} := \DX^{(\ell+1)} \phi^\prime (\mz^{(\ell)}) & 0 \le \ell \le L-1 \\
    \fDW^{(\ell)} := -\mu \DZ^{(\ell)} {\mx^{(\ell)}}^T, & 0 \le \ell \le L-1\\
    \DX^{(\ell)} := {\mw^{(\ell)}}^T \DZ^{(\ell)}, & 1 \le \ell \le
    L-1,
  \end{cases}
\end{equation}
by considering mini-batches. Algorithm \ref{alg:backward} shows the pseudocode of the backward phase for a mini-batch.
\begin{algorithm}[H]
  \caption{Backward phase for an \ac{mlp}}
  \label{alg:backward}
  \begin{algorithmic}[1]
    \Require arrays of input matrices $\mx$, weight matrices $\mw$,
    output matrices $\mz$, number of layers $L$, Loss and learning
    rate $\mu$ \State $\DX^{(L)} \gets \pd{Loss}{\mx^{(L)}}$
    \For{$\ell = L-1, L-2, \dots, 0$} \State
    $\DZ^{(\ell)} \gets \DX^{(\ell+1)} \phi^\prime(\mz^{(\ell)})$
    \State $\fDW^{(\ell)} \gets -\mu \DZ^{(\ell)} {\mx^{(\ell)}}^T$
    \If{\(\ell \ge 1\)} \State
    $\DX^{(\ell)} \gets {\mw^{(\ell)}}^T \DZ^{(\ell)}$ \EndIf \EndFor
    \State \Return $\fDW$ \Comment{Weight Changes}
  \end{algorithmic}
\end{algorithm}

\subsection{Least Mean Squares (LMS)}
LMS is an algorithm for on-line training of a single layer network with linear activation function, hence $\phi^\prime(\bullet)=1$. Consequently, the network computes \(\vz = \mw \vx\). The error of each output element of this vector is calculated by
\begin{align}
  e_{j} &= t_{j}-z_{j}, \quad for \quad 1 \leq j \leq D^{(1)} \label{error in lms}
\end{align}
which $t_j$, $z_j$ and $e_{j}$ are the $j$th elements of the ideal target, current output, and error vectors respectively. LMS computes the \ac{mse} loss as
\begin{align}
  \loss^\text{MSE}(\vz, \vt) &= \frac{1}{2} \sum_{j=1}^{D^{(1)}}{e_j}^2 =
                               \frac{1}{2} \sum_{j=1}^{D^{(1)}}{(t_{j} -
                               z_{j})}^2,
\end{align}
which yields
\begin{align}
  \pd{\loss^\text{MSE}(\vz, \vt)}{z_j} &= -(t_j-z_j), \quad for \quad 1 \leq j \leq D^{(1)}
\end{align}
or
\begin{align}
  \Dz^\text{MSE} &= \pd{\loss^\text{MSE}(\vz, \vt)}{\vz} = -(\vt-\vz) = -\ve.
\end{align}
Therefore, the derivatives vector of the loss w.r.t. the outputs is just the negative of the error vector. For $i \in \{1, 2, \dots, D^{(1)}\}$, $j \in \{1, 2, \dots, D^{(0)}\}$ and learning rate $\mu$ the chain rule can be followed to find the weight update of
\begin{align}
  \fdw_{ij}
  &:= -\mu \pd{\loss^\text{MSE}(\vz, \vt)}{z_{j}} \pd{z_{j}}{w_{i, j}}
    =\mu e_{j} x_{i}
\end{align}
or
\begin{align}
  \fDW &:= -\mu \Dz^\text{MSE} \vx^T 
         = \mu \ve \vx^T \label{eq:mu lms},
\end{align}
in the opposite direction of gradients of the loss w.r.t. weights. This is the $\mu$-LMS algorithm.

\subsection{Normalized Least Mean Squares (NLMS) \label{subsec:nlms}}
Since the inputs and targets of a single sample is all that we have at the current moment, let us foresee the consequences of applying this weight modification on the error of the current sample. By applying \eq{eq:mu lms}, the errors would be
\begin{align}
  \fDe &= -\fDz := -\fDW \vx = (\mu \ve \vx^T) \vx = \mu \ve \parallel \vx \parallel _2^2. \label{eq:correlation as squared norm}
\end{align}
As the actual correction of the error depends on the square of the second norm of the input that is not ideal. Changing the learning rate to
\begin{equation}
  \alpha := \frac{\mu}{\parallel \vx \parallel _2^2} \label{eq:adaptive learning rate}
\end{equation}
can eliminate it from the equation. After that, by applying this new learning rate \eq{eq:mu lms} becomes
\begin{equation}
  \fDW := \alpha \ve \vx^T = \frac{\mu}{\parallel \vx \parallel _2^2} \ve \vx^T, \label{eq:nlms}
\end{equation}
which is called NLMS or $\alpha$-LMS algorithm.
Finally, \eq{eq:nlms} yields
\begin{equation}
  \fDe = -\fDz := -\fDW \vx = \alpha \ve \parallel \vx \parallel _2^2
  = \mu e
  = -\mu \Dz^\text{MSE}. \label{eq:ideal error change}
\end{equation}
This particular formula has some interesting characteristics that worth extending to \ac{bp}. We discuss its properties in the next section. We also show that assuming \eq{eq:adaptive learning rate} as a learning rate is inaccurate, and there is another reasonable explanation.

\section{Proposed method \label{proposed method}}
While attention is usually only paid to the minimization of the loss function in optimization algorithms, we also pay attention to how the outputs of each layer converge. The negative direction of the gradients of the loss w.r.t. the outputs, points to some virtual targets that each layer tries to achieve. We aim to follow these imaginary targets more accurately for each sample of a mini-batch, without making drastic changes to the BP.\par
The first conclusion from \eq{eq:ideal error change} is that the error is minimized by a fraction of the previous error. It is important because just the learning rate determines the minimized error, not the specifications of the input. Another interpretation is that it updates the weights to indirectly minimize the error in the opposite direction of the gradients of the loss w.r.t. the outputs.\par
These are key takeaways from NLMS that we call them consequentialism interpretation. We want to generalize this interpretation to \ac{bp}. To sum up, we want to foresee what happens after applying the weight update formula, then to improve it by considering the consequences of our actions.\par
Our objective is first to introduce the idea of virtual targets. After that, we extend the consequentialism view to the \ac{bp} in on-line mode with the help of the virtual targets. Next, we explain why simply extending to mini-batch training would not work well, and at last, how to obtain a comprehensive formula.

\subsection{Virtual targets} \label{generalization of LMS to BP}
Here, we want to show that the on-line training of \ac{bp} with any loss function identical to training each layer of the network in isolation by $\mu$-LMS and considering virtual targets. To begin, we define
\begin{align}
  \vDz^{(\ell)} := \Dz^{(\ell)} \label{eq:virtual dz}
\end{align}
as virtual gradients, and
\begin{align}
  \tilde{\vt}^{(\ell)} := \vz^{(\ell)} - \vDz^{(\ell)} = \vz^{(\ell)} - \Dz^{(\ell)} \label{eq:virtual targets-before}
\end{align}
as virtual targets vectors for layer \(\ell\).
\begin{prop} \label{thm:virtual_targets}
  Considering elements of $\tilde{\vt}^{(\ell)}$ as virtual targets for layer $\ell$ and updating the weights in that layer with $\mu$-LMS gives us exactly the same weight update rule in \eq{eq:recursive backward} of \ac{bp} for an arbitrary loss function.
\end{prop}
\proof
From \eq{eq:recursive backward}, for layer \(\ell\) we have
\begin{align}
  \fDW^{(\ell)} := -\mu \Dz^{(\ell)} {\vx^{(\ell)}}^T
  = -\mu \pd{\mathcal{L}(\vx^{(L)},\vt)}{\vz^{(\ell)}}{\vx^{(\ell)}}^T. \label{eq:weight changes}
\end{align}

$\Dz^{(\ell)}$ is just an assigned vector of numbers at the end of the calculations. By repeating the same steps we did in \eq{error in lms} to \eq{eq:mu lms} again, we obtain

\begin{align}
  \vve^{(\ell)} &= \tilde{\vt}^{(\ell)} - \vz^{(\ell)}, \label{eq:virtual error} \\
  \loss^\text{MSE}(\vz^{(\ell)}, \vvt^{(\ell)})
                &=
                  \frac{1}{2}
                  \sum_{j=1}^{D^{(\ell+1)}}{(\tilde{t}_{j}^{(\ell)} -
                  z_{j}^{(\ell)})}^2 \Rightarrow \\
  \pd{\loss^\text{MSE}(\vz^{(\ell)}, \vvt^{(\ell)})}{\vz^{(\ell)}}
                &= -(\vvt^{(\ell)}-\vz^{(\ell)})
                  = -\vve^{(\ell)} \Rightarrow \\
  \fDW^{(\ell)} &:= - \mu \pd{\loss^\text{MSE}(\vz^{(\ell)},
                  \vvt^{(\ell)})}{\vz^{(\ell)}} \vx^{{(\ell)}^T}
                  = \mu \vve^{(\ell)} \vx^{{(\ell)}^T}. \label{eq:dw virtual targets and virutal error}
\end{align}
By substituting the real values of the virtual variables from \eq{eq:virtual error} and \eq{eq:virtual targets-before} we get
\begin{align}
  \vve^{(\ell)} &= \tilde{\vt}^{(\ell)} - \vz^{(\ell)}
                  = (\vz^{(\ell)} - \Dz^{(\ell)}) - z \Rightarrow\\
  \vve^{(\ell)} &= -\Dz^{(\ell)}, \label{eq:aliases for gradient}
\end{align}
and finally, \eq{eq:dw virtual targets and virutal error} becomes
\begin{align}
  \fDW^{(\ell)} := -\mu \Dz^{(\ell) }\vx^{{(\ell)}^T},
\end{align}
which is the same weight changes in \eq{eq:recursive backward}, and the proof is complete.
\endproof
As the current mini-batch is the only data that we have at a specific iteration, these targets are our best predictions of where the output vectors should stand in the next iteration for this very mini-batch.\par
If we manage to successfully predict and control the behavior of the outputs for each layer, since they would become the inputs of the next layer, after the non-linearities, it is possible to control the inputs of the afterward layers. Therefore, there would be no necessity to change or tune the hyper-parameters, such as learning rate, during the training.

\subsection{First extension \label{simple extension}}
As we saw in \sct{sec:past works}, the main strength of NLMS is that it minimizes the error without the input interference.\par
In our first attempt, we took a straightforward approach. We wanted to extend NLMS to feed-forward neural networks by normalizing each single input vector. Now, we show what makes this extension unstable.\par
Let us create a matrix $\hat{X}^{(\ell)}$ by normalizing each input in column $n \in \{1, 2, \dots, N\}$ of the input vector in a mini-batch as:
\begin{align}
  \bf{\hat{\vx}}_{(n)}^{(\ell)} 
  := \frac{1}{\parallel \vx_{(n)}^{(\ell)} \parallel_2^2 + \epsilon} \vx_{(n)}^{(\ell)},
\end{align}
where $\vx_{(n)}^{(\ell)}$ is the $n$-th input vector, and $\epsilon \ge 0$ is a small positive number for stability. As we saw in \sct{sec:past works}, in \ac{bp}, \eq{eq:dw virtual targets and virutal error} will be generalized to matrices of mini-batches as
\begin{align}
  \fDW^{(\ell)} := \mu \bf{\tilde{E}}^{(\ell)} \bf{\hat{X}}^{{(\ell)}^T}.
\end{align}
This extension can be refuted by a simple counterexample. Let us test it on a single layer network. By defining the matrix $\mx$ with two features and batch-size of 2 as
\begin{align}
  \mx ={
  \begin{bmatrix}
    x_{11} & x_{12}\\
    x_{21} & x_{22}
  \end{bmatrix}},
\end{align}
the $\hat{\bf{X}}$ will be
\begin{align}
  \bf{\hat{X}} = 
  \begin{bmatrix}
    \frac{x_{11}}{x_{11}^2 + x_{21}^2} &
    \frac{x_{12}}{x_{12}^2 + x_{22}^2}\\
    \frac{x_{21}}{x_{11}^2 + x_{21}^2} & \frac{x_{22}}{x_{12}^2 +
      x_{22}^2}
  \end{bmatrix}.
\end{align}
Let us look ahead by forwarding the weight changes:
\begin{align}
  \fDE := \mu \bf{\tilde{E}} \hat{X}^T \mx.
\end{align}
This yields
\begin{align}
  \fvDE := \mu \bf{\tilde{E}}
  \begin{bmatrix}
    1 &
    \frac{x_{11} x_{12} + x_{21} x_{22}}{x_{11}^2 + x_{21}^2}\\
    \frac{x_{11} x_{12} + x_{21} x_{22}}{x_{12}^2 + x_{22}^2} & 1
  \end{bmatrix},
\end{align}
where we assumed $\epsilon=0$.

The only way to move precisely towards the virtual targets happens when the right-hand side matrix is an identity matrix. It requires the input matrix to consist of orthogonal vectors, which is unlikely. In \sct{experiments}, we show what exactly happens in practice.\par
This equation shows how one sample can hinder or interfere with the optimization of another sample in a mini-batch. It also explains the mentioned bias towards stronger inputs.

\subsection{The main proposal}
Let us have a consequentialism view at \ac{bp} for a specific layer to find out where the calculations deviate from our expectations. In mini-batch mode, from \eq{eq:dw virtual targets and virutal error} and \eq{eq:aliases for gradient}, we have
\begin{align}
  \fDW^{(\ell)} := -\mu \DZ^{(\ell)} \mx^{{(\ell)}^T} = \mu \vme^{(\ell)} \mx^{{(\ell)}^T}. \label{original update rule-the matrix representation}
\end{align}
Now, we want to find out what exactly happens at the next iteration if we see the same mini-batch again. From \eq{forward formula with step} we have
\begin{align}
  \fvDZ^{(\ell)} = \fDW^{(\ell)} X^{(\ell)}. \label{eq:real weight changes in forward}
\end{align}
By substituting $\fDW^{(\ell)}$ from \eq{original update rule-the matrix representation} in the above equation, we get the real changes in the error matrix:
\begin{align}
  \fvDE^{(\ell)} := \mu \vme^{(\ell)} (\mx^{{(\ell)}^T} \mx^{(\ell)}). \label{how do we get the correlation matrix}
\end{align}
It is where a correlation matrix emerges. Thus, $\parallel \mx \parallel_2^2$ in equation \eq{eq:correlation as squared norm}, in $\mu$-LMS, becomes $\mx^{{(\ell)}^T} \mx^{(\ell)}$ for a mini-batch in \ac{bp}. It is the reason why the loss surface bends sharply in some dimensions than the others.\par
We expect \eq{how do we get the correlation matrix} to deflect the outputs from the desired virtual targets. This redundant calculation can confuse \ac{dnn}s. In \sct{experiments}, by a toy example, we show what exactly happens.\par
The next step is to devise an analytical solution. The idea is to modify the weight correction rule in such a way that produces an identity matrix at the next hypothetical iteration instead of the mentioned correlation matrix. Therefore, we extend \eq{eq:ideal error change} from the NLMS algorithm to mini-batch training and solve it.\par
We wish to have $\fvDE$ matrix as our desired error reduction to be
\begin{align}
  \fvDE^{(\ell)} = \mu \vme^{(\ell)}.
\end{align}
The solution to
\begin{align}
  \bf{\vec{\Delta} \widehat{W}^{(\ell)}} \mx^{(\ell)} = \mu \vme^{(\ell)},
\end{align}
is our answer, where $\bf{\vec{\Delta} \widehat{W}}$ is our ideal matrix of weight changes. By solving it, we find our desired weight update rule as
\begin{align}
  \bf{\vec{\Delta} \widehat{W}^{(\ell)}} := \mu \vme^{(\ell)}
  \mx^{{(\ell)}^\dagger}= -\mu \DZ^{(\ell)} \mx^{{(\ell)}^\dagger}, \label{ideal weight update rule}
\end{align}
where the $\dagger$ is the pseudoinverse operator. To ensure that it works as expected, we repeat the substitution of $\fDW$ in \eq{eq:real weight changes in forward} by this new weight update rule of $\bf{\vec{\Delta} \widehat{W}}$ we get
\begin{align}
  \fvDE^{(\ell)} &:= \mu \vme^{(\ell)} \left(\mx^{{(\ell)}^\dagger} \mx^{(\ell)}\right) \\
                 &\approx \mu \vme^{(\ell)} \bf{I}^{(\ell)} = \mu \vme^{(\ell)},
\end{align}
where $\bf{I}^{(\ell)}$ is the identity matrix. Since the pseudoinverse gives us the best solution based on least squares solution, the outputs of each layer should move to the best possible direction of the virtual targets in current iteration.\par
The perfect learning rate for a single layer with a linear activation function in full-batch mode should be $\mu=1$. However, as we explain, it needs a small but important final modification.\par

Since the partial derivatives are calculated as limits to find the tangent hyperplane to optimize a complicated function, we are eventually bound to use numerical approximations in practice. Thus, there is an implicit assumption that when we can change all of them at the same time, the mentioned assumption holds.\par

We consider all layers as isolated layers for small variable changes. This assumption may also be violated if we change weights drastically. Therefore, we have to not only take care of the weight changes for a specific layer but also for the other variables on the next layers that are also being affected by the weight changes in the previous layers.\par
We use the Moore-Penrose method with ridge regression \cite[pp. 45-46]{Goodfellow-et-al-2016} \cite{hoerl1970ridge} for two purposes. In the first place, it can easily prevent ill-conditioning issues. Besides, it also prevents the weights from changing dramatically, anticipating the violation of the chain rule. Therefore, \eq{ideal weight update rule} can be rewritten as
\begin{equation}
  \bf{\vec{\Delta} \widehat{W}^{(\ell)}} := 
  -\mu \DZ (\mx^{{(\ell)}^T} \mx^{(\ell)} 
  + \lambda \bf{I}^{(\ell)})^{-1} \mx^{{(\ell)}^T}. \label{eq:consequentialism formula}
\end{equation}
We refer to the complete formula of \eq{eq:consequentialism formula} as the consequentialism weight update rule. The pseudocode of the proposed algorithm is shown in Algorithm \ref{alg:backward consequentialism}.\par
\begin{figure}
  \begin{algorithm}[H]
    \caption{Backward phase for a feed-forward network with
      consequentialism weight updates}
    \label{alg:backward consequentialism}
    \begin{algorithmic}[1]
      \Require arrays of input matrices $\mx$, weight matrices $\mx$,
      output matrices $\mz$, number of layers $L$, Loss and learning
      rate $\mu$ \State $\DX^{(L)} \gets \pd{Loss}{\mx^{(L)}}$
      \For{$\ell = L-1, L-2, \dots, 0$} \State
      $\DZ^{(\ell)} \gets \DX^{(\ell+1)} \phi^\prime(\mz^{(\ell)})$
      \State Compute the ${\mx^{(\ell)}}^\dagger$ with ridge
      regression \State
      $\fDW^{(\ell)} \gets -\mu \DZ^{(\ell)} {\mx^{(\ell)}}^\dagger$
      \If{\(\ell \ge 1\)} \State
      $\DX^{(\ell)} \gets {\mw^{(\ell)}}^T \DZ^{(\ell)}$ \EndIf
      \EndFor \State \Return $\fDW^{(\ell)}$ \Comment{Weight changes}
    \end{algorithmic}
  \end{algorithm}
\end{figure}
The new method applies to all types of layers as they are, in fact, some particular cases of the matrix multiplication or FC layer. For example, the convolution operator almost always is implemented as a matrix multiplication by using the im2col method for forward and col2im method for backward \cite{jia2014caffe}.\par

\subsection{Consequentialism is not normalization}
Our method is not a normalization algorithm. Hence, one should not expect the behaviors of a normalization algorithm. However, in some ways, its objectives may overlap with those normalization algorithms. For instance, it cancels the effect of input on weight updates that is similar to what a normalization algorithm offer, but it does not change the inputs of the layers as we see in BN.\par
In the following, we show that even NLMS, as a particular form of our generalized method, is not a normalization algorithm. Let us rewrite the pseudoinverse of the input vector in NLMS as
\begin{align}
  \vx^\dagger &= \frac{1}{\parallel \vx \parallel_2^2} \vx^T =
                \frac{1}{\parallel \vx \parallel_2} \frac{\vx^T}{\parallel \vx
                \parallel_2} = \frac{1}{\parallel \vx \parallel_2} \bf{\hat{x}}, \label{eq:insight}
\end{align}
where \(\bf{\hat{x}}\) is the normalized input vector (unit vector) of \(\vx\).\par
It explains that the \(\frac{\vx^T}{\parallel \vx \parallel_2^2}\) part of the NLMS should be considered as a pseudoinverse of a vector, instead of considering \(\frac{\mu}{\parallel \vx \parallel_2^2}\) an adaptive learning rate. It has some interesting outcomes.\par
It indicates how the norm of the input affects the optimization twice. First, on the backward pass, then on the next forward pass. It means we are dividing the unit vector of input by the second norm to neutralize the effect of input power in the weight update calculations.\par
By considering a single scalar input as the simplest form, we see that we are dividing the gradient by the input, not multiplying, to neutralize the multiplication to the input that we have in the next hypothetical forward pass.\par

\section{Experiments \label{experiments}}
In this section, we evaluate the proposed algorithm by several criteria: loss surface of parameters and the convergence path of the outputs to the targets on a toy example, learning curves of an \ac{mlp} on multiple datasets, and training of ResNet \cite{He2015DelvingDI} with different optimizers.\par
We used a 3 GHz Pentium G2030 CPU and a GTX 750 2GB GPU for our experiments. \ac{mlp} experiments were carried out in Tensorflow \cite{tensorflow2015-whitepaper} and ResNet experiments were performed in Caffe \cite{jia2014caffe}.\par
Here, we should clarify that we had to disable cuDNN \cite{chetlur2014cudnn} in Caffe to use the im2col function. It had an adverse effect that we could not utilize the optimization and GPU processing that cuDNN brings. Therefore, our method may seem slower in ResNet tests, which is implemented in Caffe, in comparison to the \ac{mlp} implementation in Tensorflow that utilized cuDNN. Besides, for a fair comparison, it is possible to use multiple GPUs to train a network asynchronously with the proposed algorithm in practice. Therefore, the computation time of pseudoinverse would be negligible as one can utilize the idle resources in the forward and the backward passes of the next layers.\par
At last, we should also mention that in this section, the "C-" prefix is used to refer to the consequentialism version of optimization algorithms. For instance, C-SGD is the consequentialism version of the SGD algorithm.

\subsection{Loss surface}
\begin{figure}[!t]
  \centering \includegraphics[width=2.3in]{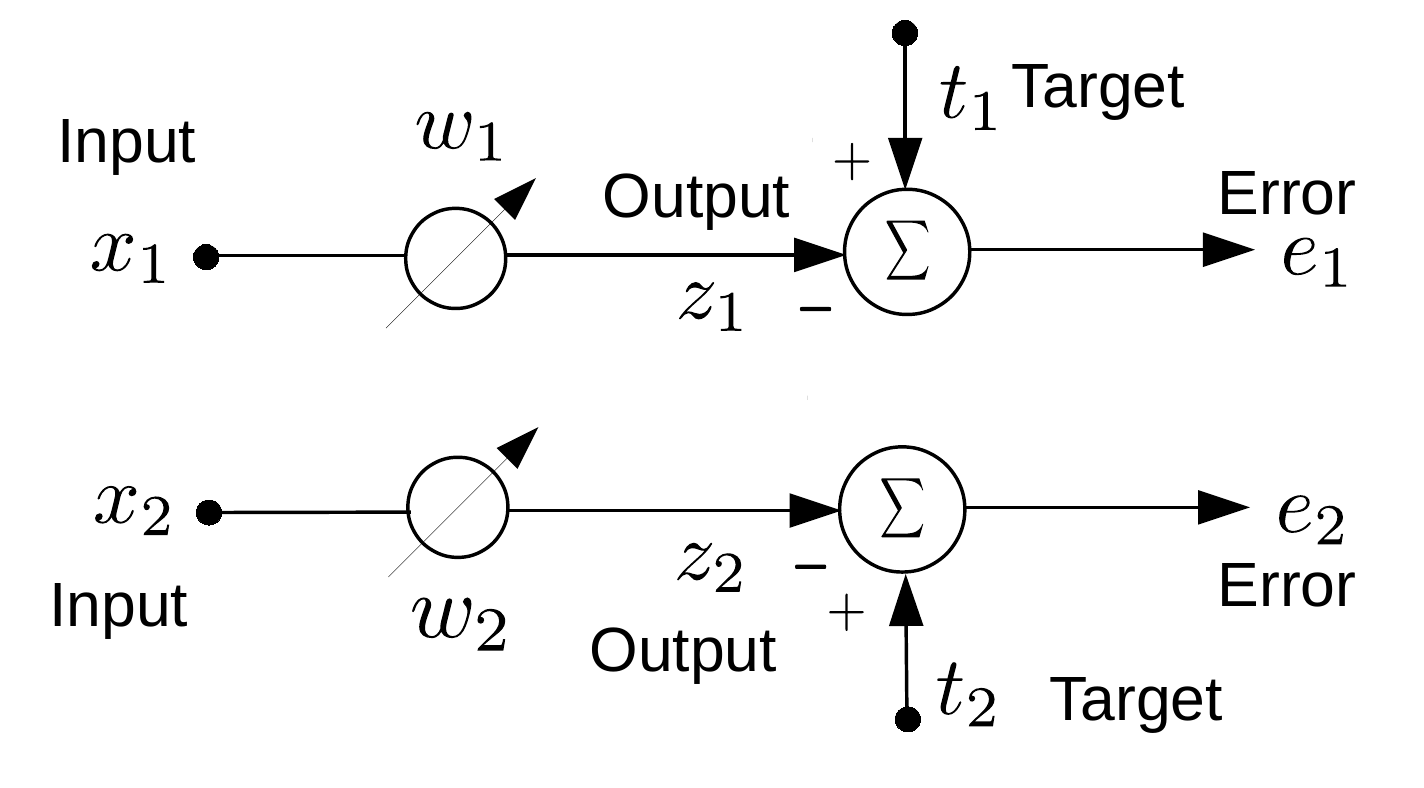}
  \caption{A simple network of two distinct inputs, weights and
    outputs.}
  \label{fig:very simple net}
\end{figure}

\begin{figure}[!b]
  \centering \includegraphics[width=3.5in]{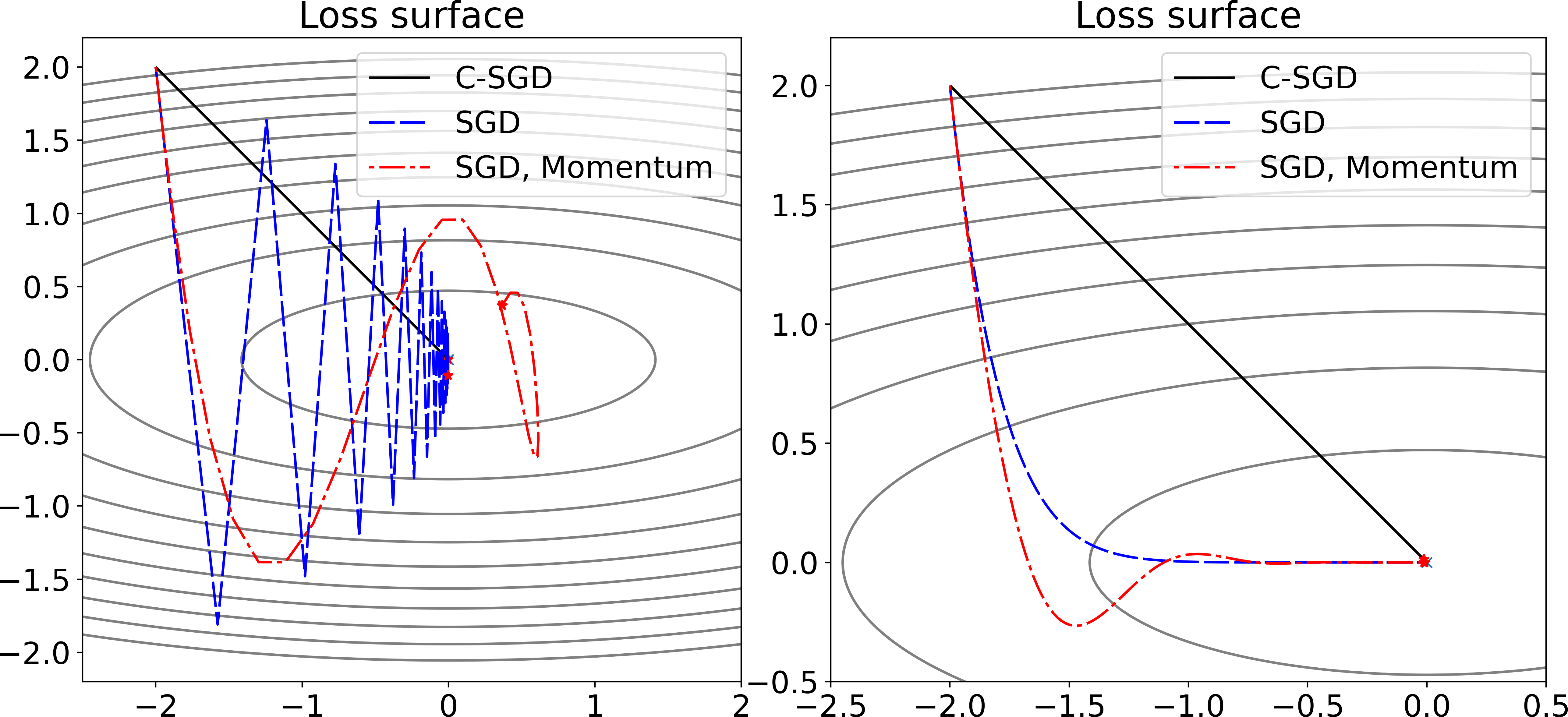}
  \caption{Loss surface for C-SGD, \ac{sgd} and \ac{sgd} with
    momentum. The left figure shows the convergence with a high
    learning rate. The right figure shows the convergence with a
    small learning rate.}
  \label{fig:loss surface}
\end{figure}

In our first experiment, we wanted to know how the weights converge to their optimum values in the loss surface graph. For this purpose, we created a very simple single-layer network with just two separate inputs and outputs with \ac{mse} loss function, which is depicted in \fig{fig:very simple net}. Also, the loss surface of two weight parameters and the path they travel to the optimum values with different learning rates are also shown in \fig{fig:loss surface}.\par

As the graphs of \fig{fig:loss surface} show, the proposed algorithm moves the parameters directly towards the optimum values, regardless of the learning rate. By contrast, the parameters move in the direction of the gradients in SGD, which is perpendicular to the ellipses in the loss surface. While a weight parameter in SGD converges faster in one direction, it progresses slowly in another one. It is why raising the learning rate to achieve faster convergence in one axis may cause some perturbations in the other one.\par
For a higher learning rate in the left graph, the momentum term reduced the oscillations of SGD in vertical axis, but the accumulated momentum caused the other parameter to pass its optimum value in horizontal axis. Since the momentum term is usually being used alongside the higher learning rates, it also can be an unexpected source of randomness that makes the selection of hyper-parameters harder.\par
In the right graph, applying a small learning rate eliminated the oscillations of the vanilla SGD and reduced them for the SGD with momentum at the cost of more iterations and wasting more time and resources.\par
To conclude, as the consequentialism makes the outputs of each layer more predictable, we expect fewer perturbations and easy hyper-parameter selection in more complex models.

\subsection{The path from the initial outputs to the ideal targets}
In our second experiment, we compared our algorithm to SGD in mini-batch mode regarding the convergence of the outputs to the targets.\par
\begin{figure}[!t]
  \centering \includegraphics[width=3.0in]{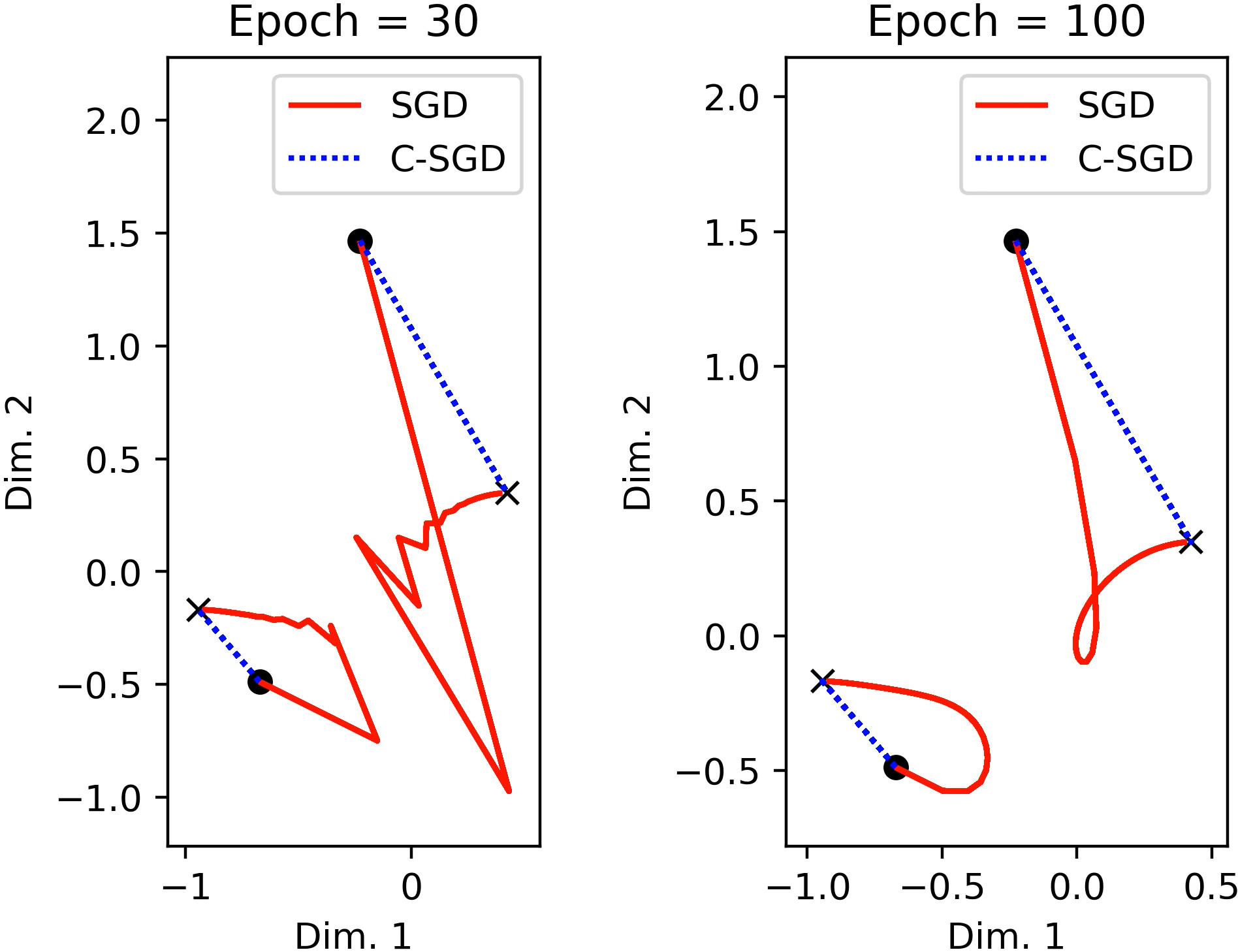}
  \caption{The convergence of the outputs of a single-layer FC network for 2 out of 10 samples of 20 dimensions inputs during the training. Outputs have 2 dimensions. The outputs start from the $\bullet$ point and should reach to the targets marked with $\times$. The learning rate for C-SGD was 0.7 for both figures. The learning rate of SGD was 0.03 for the left graph and 0.01 for the right graph.}
  \label{outputs paths}
\end{figure}

The network in this experiment has only a single FC layer that produces 2-dimension outputs from 20-dimension inputs. All samples are random numbers from a normal distribution, and the weights initialized with Xavier initialization \cite{Glorot2010UnderstandingTD}. \fig{outputs paths} depicts the convergence path of the outputs of 2 out of 10 samples to their targets for both SGD and C-SGD.\par
As the graphs illustrate, regardless of the learning rate, the proposed method steers the outputs directly towards the targets. Conversely, SGD not only took an indirect path but also demonstrated more oscillations. While this usually happens in practice with higher learning rates, as we see, the mini-batch training also increases the unpredictability.\par
The right graph shows that even decreasing the learning rate does not eliminate the fluctuations, because one or more of the input samples can affect the other samples from behaving predictably. Consequently, the learning rate is not the only cause of the observed turbulences in the mini-batch training of the SGD algorithm.\par
To sum up, these plots show how the combination of increasing the learning rate and deploying the mini-batch training for SGD can cause the outputs of a layer, which will be the inputs of the next layer after a non-linearity in \ac{dnn}s, to oscillate around the optimum values (virtual targets) while the proposed algorithm shows stability in all cases.

\subsection{Fully-connected neural networks}
Here, we investigate the effectiveness of the consequentialism version of the SGD against the plain SGD on a single \ac{mlp} architecture for three datasets in Tensorflow that is one of the most popular machine learning libraries. As we will see in the more comprehensive experiments of the next subsection, SGD does better than other optimizers such as Adam and Nesterov. Accordingly, we limited our experiments in this subsection just to SGD.\par
All networks in the following experiments had eight hidden layers, and the batch size was 32. Moreover, weights were initialized by Kaiming's method \cite{He2015DelvingDI}, all layers except the last one had \ac{relu} activation functions, the last layer had a softmax activation, and we applied the cross-entropy loss. We used the full training set to report the loss.\par
In these experiments, each layer of the networks for CIFAR-10 \cite{krizhevsky2009learning} and Fashion-MNIST \cite{xiao2017/online} had 800 neurons. The network used for CIFAR-100 \cite{krizhevsky2009learning} had 1600 neurons. For half of the experiments, the inputs of the first layer were normalized to be between 0 and 1, and $\lambda$ was set to $1e-3$ for the consequentialism formula. For the other half, each feature was normalized to zero mean and unit variance (marked by NF), and $\lambda$ was set to \(1e-2\).\par
\begin{figure*}[!tb]
  \centering
  \includegraphics[width=0.8\linewidth]{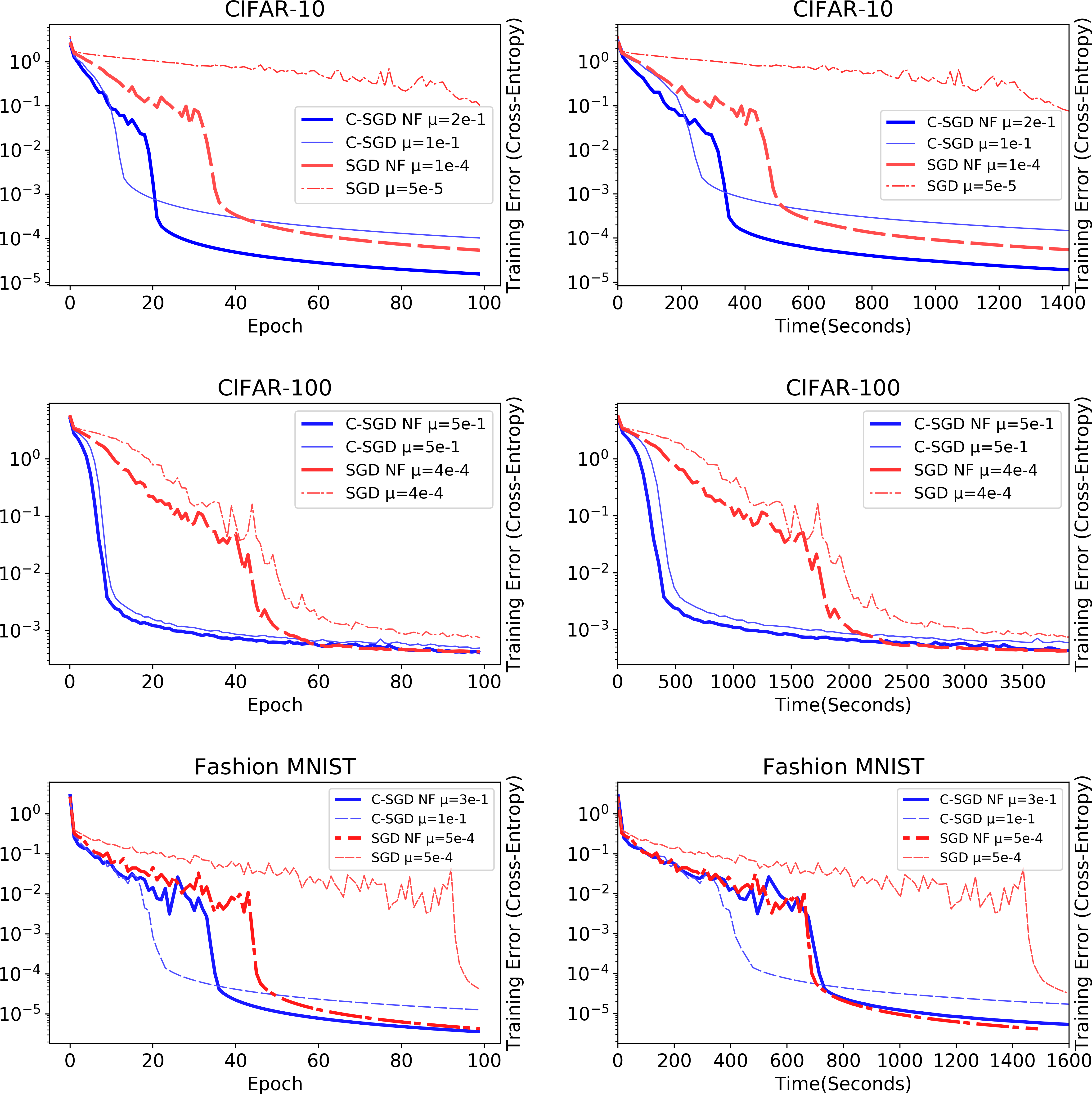}
  \caption{Log scaled cross-entropy loss of an 8-layer FC neural network with \ac{relu} activations, trained with \ac{sgd} and C-SGD on different datasets. NF denotes normalized features. The left graphs display the loss for all training samples at each iteration, and the right graphs display the same values based on the elapsed time.}
  \label{training loss for MLP}
\end{figure*}

\fig{training loss for MLP} shows the loss graph of SGD against C-SGD over 100 epochs for all datasets based on iterations and elapsed time.\par
According to these experiments, C-SGD optimizes faster than SGD from the beginning regardless of the distribution of the features. Moreover, the graphs display more overall stability throughout the training for C-SGD. Since SGD is more sensitive to the probability distribution of the inputs, it suffers more from the wrongly chosen distribution or some unexpected changes during the training.\par
The graphs also show that both algorithms reach a point that the loss drops sharply, then the progress becomes slow, which usually occurs earlier for C-SGD. It also implies C-SGD takes the more directed path towards virtual targets.\par
Since C-SGD needs more time in single GPU training because of the pseudoinverse calculations, we also compared them by elapsed time. The right-hand side graphs of \fig{training loss for MLP} depict the comparison between both algorithms in seconds.\par
According to these experiments, C-SGD maintains its competitiveness for CIFAR-10 and CIFAR-100 even by taking the time into account. For Fashion-MNIST, their performances are almost identical. It may happen due to the more preprocessing that was performed on this dataset that the pseudoinverse does not help a lot. For example, the images have not much skewing, rotations, and significant portions of the images are just black, but the improvement is substantial for a more complicated and more intact dataset. Therefore, the loss for CIFAR-100 drops faster than CIFAR-10.\par
As a final observation, for more complicated inputs of the CIFAR-100, the feature normalization does help neither.
\subsection{ResNet-20}
In this subsection, we study the effectiveness of the proposed method on the deeper architecture of the ResNet-20 \cite{He2015DelvingDI}. We applied our method to both FC and convolutional layers in Caffe \cite{jia2014caffe}.\par
Since the number of combinations of optimizers, models, datasets was quite large, we limited our tests to the CIFAR-10 dataset and optimizers to SGD \cite{rumelhart_learning_1986}, Nesterov \cite{ruder2016overview}, and Adam \cite{Kingma2014AdamAM}, among the popular optimizers.\par
As the main idea of the consequentialism and BN was canceling the confusing effect of inputs on optimization as a significant source of the unpredictability in the neural networks, we observed the combination of them did not help, and even they had some conflicts. Consequently, we removed the BN layers for the consequentialism algorithm in all of our tests. However, we tested all other optimizers with and without BN layers.\par
These were our settings: $\lambda$ was 0.03, and batch-size was 128. The momentum value of 0.95 for both SGD and Nesterov showed the best results in our experiments. We used the default values of 0.9 and 0.999 in \cite{Kingma2014AdamAM} for the first and second moments for Adam. The inputs were normalized between -1 and 1 for plain optimizers, and between -0.5 and 0.5 for the consequentialism method. The learning rate multiplier of the FC layer was 0.1 for the consequentialism experiments. The next graphs show the values from the last mini-batch of each 100 iterations as reported by Caffe.\par
\begin{figure*}[!tbp]
  \centering
  \includegraphics[width=0.9\linewidth]{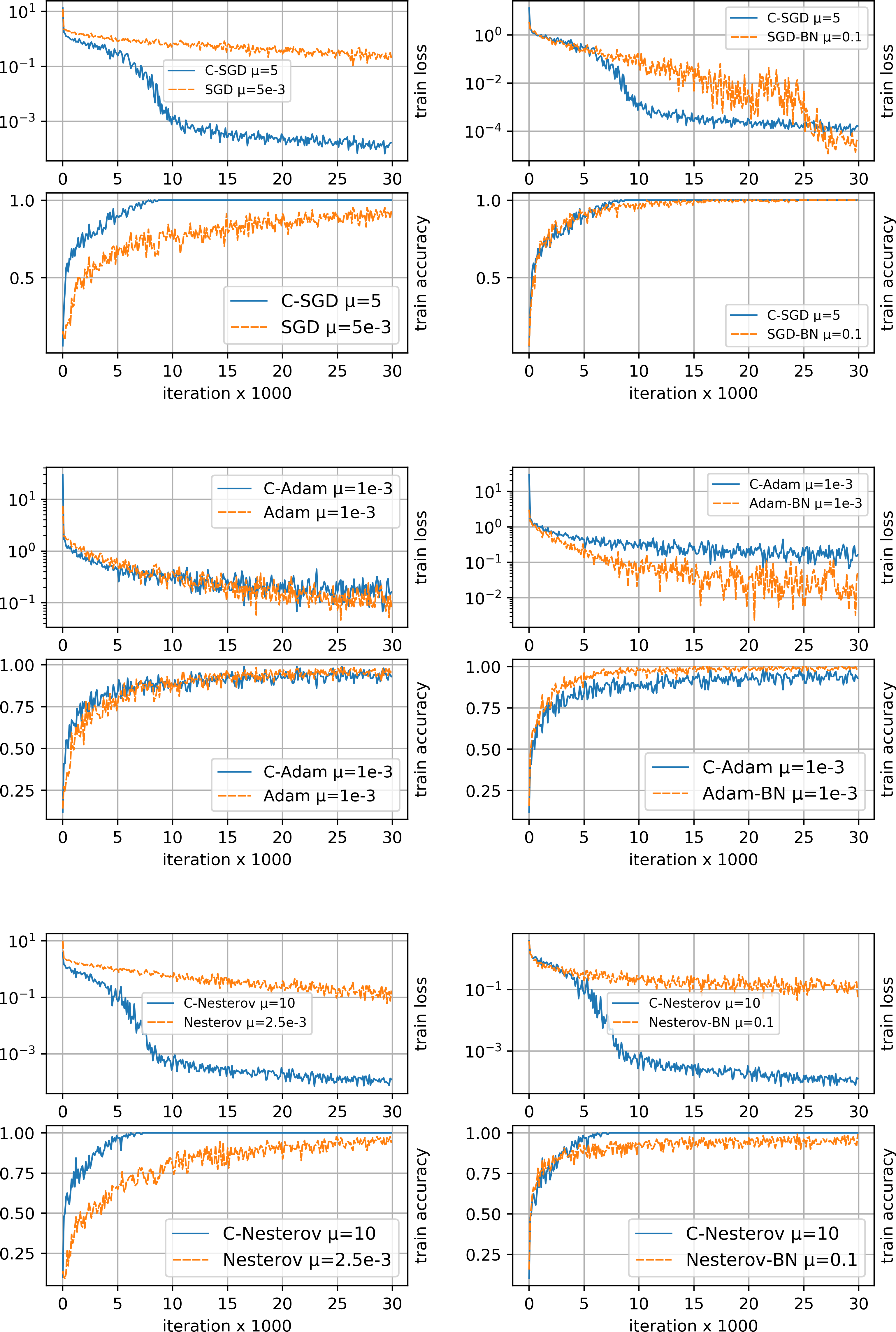}
  \caption{Comparison of log scaled cross-entropy loss and accuracy of ResNet-20 on CIFAR-10 trained by various optimizers with and without BN layers. BN denotes the models with BN layers.}
  \label{resnet-20}
\end{figure*}
\fig{resnet-20} compares the loss and accuracy of the consequentialism form of optimizers alongside their plain counterparts for two different architectures, first in the presence, then in the absence of the BN layers.\par
\begin{figure*}[!tbp]
  \centering
  \includegraphics[width=0.9\linewidth]{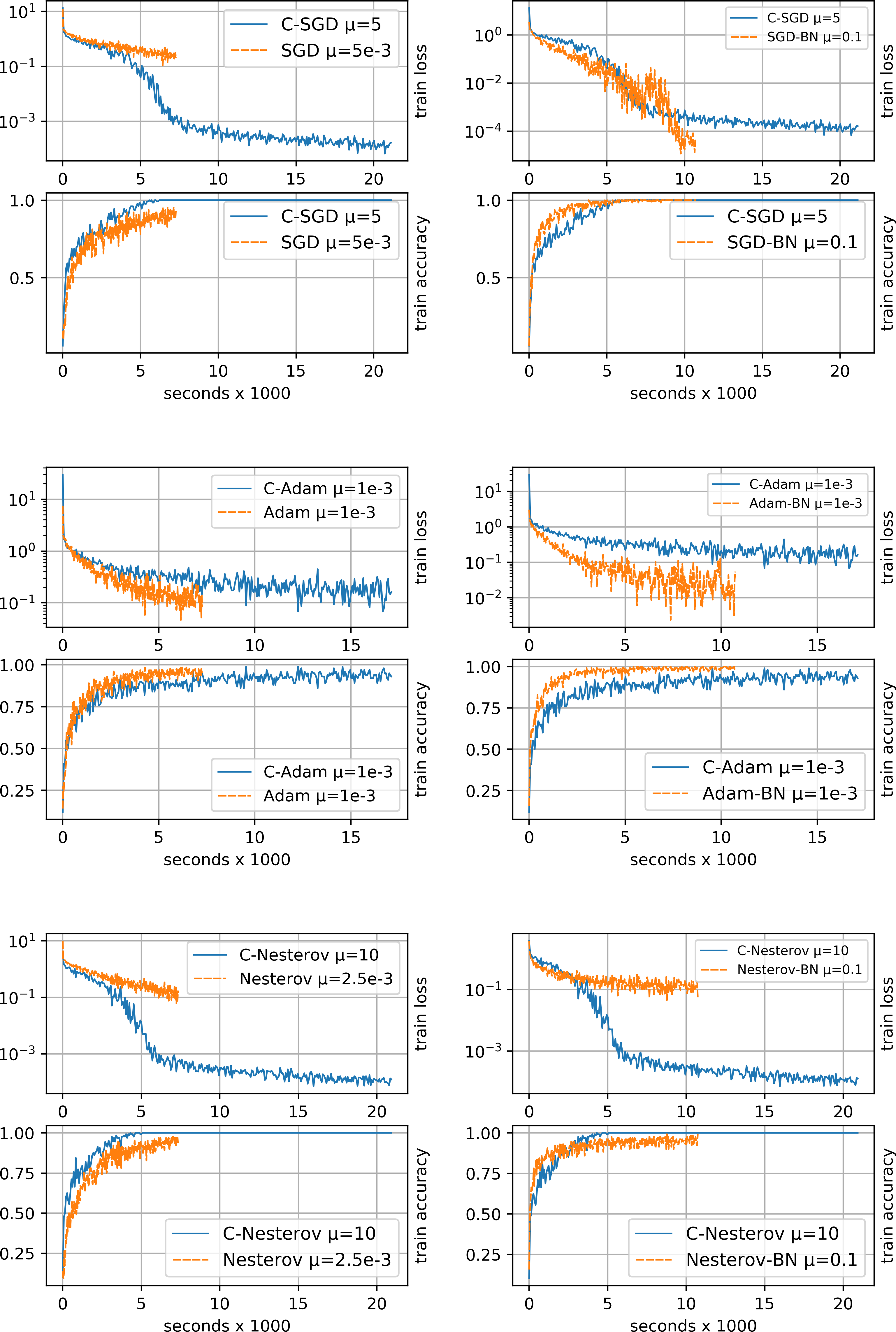}
  \caption{Time-based comparison of log scaled cross-entropy loss and accuracy of ResNet-20 on CIFAR-10 trained by various optimizers with and without BN layers. BN denotes the models with BN layers.
  }
  \label{time-based resnet-20}
\end{figure*}
For the right-hand side graphs, we deployed the unmodified models with BN layers, and for the left-hand side graphs, we removed the BN layers. As the non-BN graphs display, C-SGD and C-Nesterov perform better than SGD and Nesterov respectively. Adam optimizer did not perform better than the other two optimizers at all, and C-Adam did not change the outcome.\par
On the other hand, while C-SGD achieved 100\% train accuracy without forgetting any sample after about 8000 iterations, the SGD with BN, sometimes forgot some samples. Overall, C-SGD converged faster in the beginning and maintained its stability to the end of the training, but SGD with BN finally crossed the C-SGD line after 25000 iterations.\par
Another interesting observation is the striking resemblance between their accuracy graphs. The above tests show that our method, like BN, does not suffer from the vanishing and exploding gradients, although each algorithm took a different path towards virtual targets. For the Nesterov, although applying the BN layers improved the convergence, C-Nesterov maintained its competitiveness and reached 100\% accuracy at about 7000 iterations despite having significantly less learnable parameters.\par
Now, let us consider the time instead of steps. \fig{time-based resnet-20} compares the loss and accuracy of the consequentialism method alongside their plain counterpart optimizers for two different architectures, in the presence and absence of the BN layers in seconds.\par
According to the left-hand side graphs, SGD and Nesterov performed better than Adam. C-SGD and C-Nesterov performed much better than their plain counterparts. However, BatchNorm versions of SGD performed almost the same by considering the time. Besides, C-Nesterov also did better than Nesterov-BN.\par
In conclusion, consequentialism can help to reach to highest accuracies in less time. Moreover, because of its more stable nature and better-chosen path to the virtual targets, it forgets the previous mini-batches less often. Therefore, it has the potential to learns faster if it is combined with a suitable regularization method.

\section{Conclusion and Future works \label{sec:conclusion}}
In this paper, we extended Normalized least mean squares (NLMS) and \acf{apa} to \acf{bp} from a new perspective. Firstly, we introduced a new way of improving the weight correction formula in BP by anticipating the consequences of our actions in the next iteration. We explained that it is equivalent to minimizing the loss with gradients of the outputs by the weight updates instead of directly considering gradients of the loss w.r.t. weights. Secondly, we showed why one can deal with the gradients of the outputs of each layer as the changes to be made towards virtual targets instead of just a mere minimization of a loss function.\par
We also proved that optimizing by BP is equivalent to optimizing each layer independently by the LMS algorithm and virtual targets. Therefore, it is possible to apply successful algorithms of adaptive filters to \acf{dnn}s.

\clearpage
\bibliography{References}

\end{document}